\documentclass{article}

\usepackage[english]{babel}

\usepackage[letterpaper,top=2cm,bottom=2cm,left=3cm,right=3cm,marginparwidth=1.75cm]{geometry}

\usepackage{amsmath}
\usepackage{graphicx}
\usepackage{authblk}
\usepackage{subfigure}
\usepackage{booktabs} 
\usepackage[colorlinks=true, allcolors=blue]{hyperref}
\usepackage{fancyhdr}

\fancypagestyle{headernote}{
  \fancyhf{} 
  \fancyhead[C]{\footnotesize\textit{THIS TECHNICAL REPORT PRESENTS SOME PRELIMINARY RESULTS OF THE RESEARCH CONDUCTED ON THIS TOPIC}} 
}
\title{Automatic counting and classification of mosquito eggs in field traps}

\author[1]{Javier Naranjo-Alcazar}
\author[1]{Jordi Grau-Haro}
\author[1]{Pedro Zuccarello}
\author[2]{David Almenar}
\author[3]{Jesus Lopez-Ballester}
\affil[1]{Instituto Tecnológico de Informática (ITI), Paterna, 46980, Valencia, Spain}
\affil[2]{Empresa de Transformación Agraria S.A., S.M.E., M.P. (TRAGSA), Paterna, Spain}
\affil[3]{Department of Computer Science, ETSE, University of Valencia, Valencia, Spain}

\begin{document}

\maketitle
\thispagestyle{headernote} 

\begin{abstract}
Insect pest control poses a global challenge, affecting public health, food safety, and the environment. Diseases like dengue, malaria, and Zika, transmitted by mosquitoes, are expanding beyond tropical regions due to climate change. Agricultural pests further exacerbate economic losses by damaging crops. The Sterile Insect Technique (SIT) emerges as a promising, eco-friendly alternative to chemical pesticides, involving the sterilization and release of male insects to curb population growth. This work focuses on the automation of the analysis of field ovitraps used to follow-up a SIT program for the \textit{Aedes albopictus} mosquito in the Valencian Community, Spain, led and funded by the Conselleria de Agricultura, Ganadería y Pesca. Previous research has leveraged deep learning algorithms to automate egg counting in ovitraps, yet faced challenges such as manual handling and limited analysis capacity. Innovations in our study include classifying eggs as hatched or unhatched and reconstructing ovitraps from partial images, mitigating issues of duplicity and cut eggs. Additionally, our device can analyze multiple ovitraps simultaneously without the need of manual replacement. This approach significantly enhances the accuracy and efficiency of egg counting and classification, providing a valuable tool for large-scale field studies and improving the monitoring of SIT programs.


\end{abstract}

\section{Introduction}

Insect pest control is a global challenge affecting public health, food safety and the natural environment. Mosquito-borne diseases, such as dengue, malaria or Zika virus, pose a significant threat to the health of the world's population. Although, traditionally, certain species of mosquitoes that act as disease vectors have been concentrated in tropical or subtropical regions, today, due to factors such as climate change, these insects have expanded their presence to geographic regions where they were not previously present \cite{naddaf2023dengue}. On the other hand, insect pests related to agricultural activity can cause significant economic losses by destroying crops and reducing food production \cite{Sharma2017}.

In this context, the Sterile Insect Technique (SIT) \cite{dyck2021sterile} is considered a promising strategy for pest control, offering a sustainable and environmentally friendly alternative to other pest control methods, such as chemical pesticides. SIT is based on mass rearing of the insect in question inside a biofactory, separation by sex (or sex-sorting), sterilization of the males, and subsequent release of these into the environment. The standard sterilization procedure is usually based on ionizing radiation. The main idea is that copulation between sterile and wild insects leads to non-viable eggs, therefore, achieving the reduction of the population of the insect to be controlled. 

\begin{figure}
    \centering
    \includegraphics[width=1\linewidth]{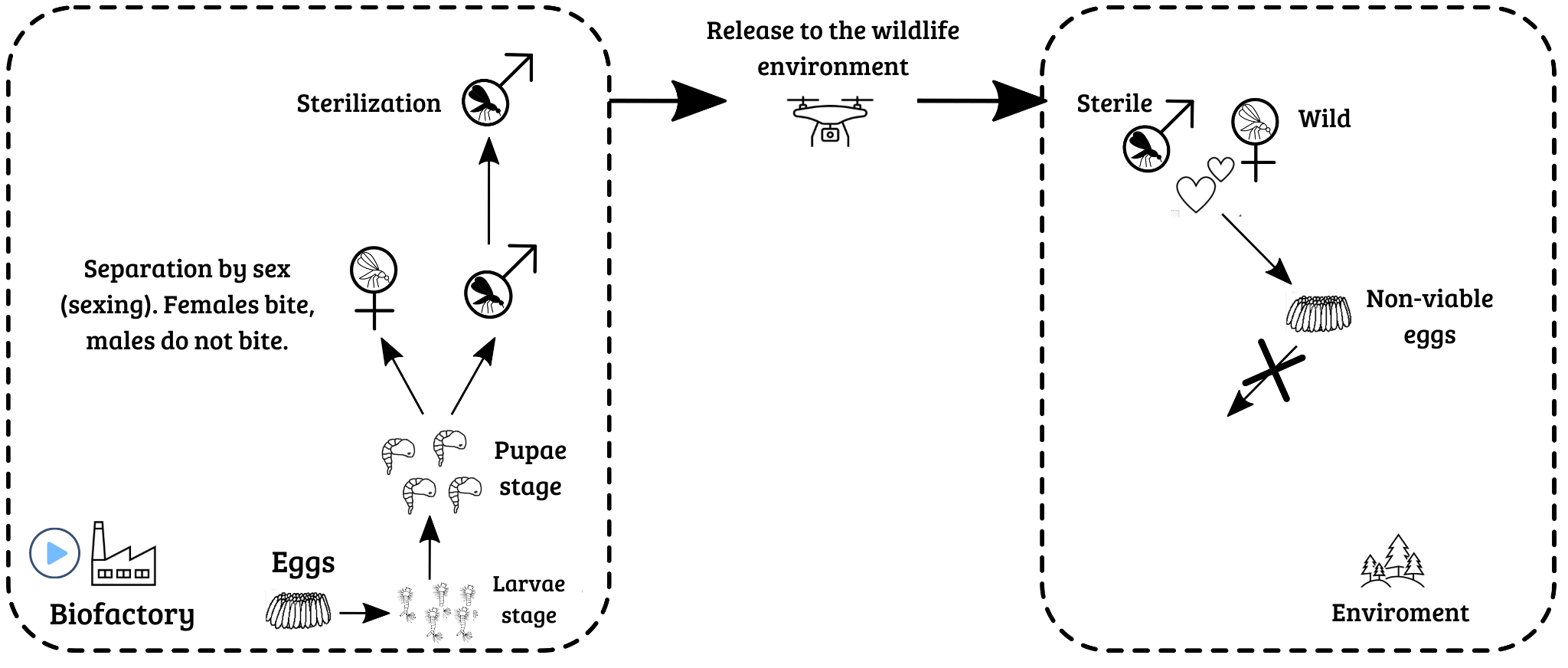}
    \caption{Infographic describing the case of the SIT of the \textit{Aedes albopictus} mosquito studied in this paper. The left rectangle shows schematically the process inside the biofactory, while the right rectangle shows what is expected to happen after the release of the sterile males into the environment.}
    \label{fig:sit_image}
\end{figure}

Within this work, the target insect is the \textit{Aedes albopictus} species of mosquito, commonly known as the tiger mosquito. The complete SIT process for this case is schematically described in Fig.~\ref{fig:sit_image} \cite{Tur2021_insects}. The release of female specimens is avoided because, unlike males, it is the females that bite humans, with the associated risk of disease transmission.  In the Valencian Community, Spain, an ongoing SIT program against Aedes albopictus, led and funded by the Conselleria de Agricultura, Ganadería y Pesca, has shown promising results in reducing the population of this invasive mosquito species \cite{Tur2021_insects}.

Once the sterile insects have been released into the environment, it is common to carry out follow-up actions to verify the incidence of the treatment on the mosquito population in the area under study. Field traps are distributed at specific locations throughout the area for the females to lay their eggs. These traps are commonly called ovitraps \cite{FayPerry1965}. After a few days, the ovitraps are collected and studied in the laboratory. The ratio of hatched/unhatched eggs observed in the traps is a figure of merit that can be used to measure the success of the SIT program, compared to this ratio in a control untreated area.


The analysis in the laboratory of these ovitraps, when done manually, is costly, both in time and personnel. In recent years, several scientific papers related to the possibilities of automating the analysis and counting of eggs in ovitraps using deep learning algorithms and models have been published, achieving significant advances in this regard. In Garcia et al. \cite{Garcia2019SIBGRAPI}, a Canon PowerShot A650 IS camera coupled to a microscope is used to acquire ovitraps images in the laboratory, processing approximately 14,500 eggs with 91\% accuracy using a decision tree-based classifier and a convolutional neural network (CNN). Although high accuracy was achieved, the main disadvantage lies in the need for manual manipulation of the ovitraps for image capture. On the other hand, de Santana et al. \cite{Santana2019IEEE} presented a fully automated device that relies on a portable microscope, raspberry-pi hardware with two motors that move the ovitrap in front of the microscope and cloud processing to analyze ovitrap images. With 91\% accuracy, this method also uses a CNN, the R-FCN \cite{Dai2016}, and offers the advantage of automatic microscope positioning. However, analysis is limited to one ovitrap at a time and requires manual repositioning, which can be a drawback in large-scale studies. Oliveira Vicente et al. \cite{OliveiraVicente2024arxiv} introduced a new dataset that includes both field and laboratory ovitrap images and performed tests with three different CNNs: Faster R-CNN \cite{Ren2015_IEEE}, SABL \cite{SABL2020} and FoveaBox \cite{Kong2020_FOVEABOX}. Although they did not find significant differences between the networks for field imaging, FoveaBox showed better results under laboratory conditions. Despite this, the study did not consider the use of different zoom levels to reduce egg overlap, which could improve performance metrics. In the conclusions, it is suggested that several new CNNs are studied and tested in the case of overcrowded ovitraps with a high density of overlapping eggs \cite{Cheng2022,Song2021,Tran2022}. Finally, Javed et al. \cite{javed2023eggcountai} used MaskRCNN \cite{He_2017_ICCV} to classify eggs in images at different zoom levels, achieving accuracies of 98.8\% for micro (high zoom) and 96.06\% for macro (low zoom) images. Although this approach showed high accuracy, no reconstruction of the complete ovitrap was performed, which can result in duplicities or cut eggs in the analysis of complete ovitraps.

The new study presented in this paper proposes significant innovations by classifying the eggs into two categories: hatched and unhatched, in contrast to previous works that did not consider egg condition. In addition, a strategy is introduced to reconstruct the ovitrap from partial images with a certain percentage of overlap, eliminating duplicities and cut eggs. In addition, the device implemented for image capture allows the analysis of several ovitraps at once without the need to replace them one by one. Overall, the approach adopted will not only improve the accuracy of egg counting and classification, but also optimize the process by allowing the analysis of complete ovitraps, providing a particularly useful tool for field studies related to SIT monitoring.




\section{Materials and Methods}\label{sec:method}

\subsection{Experimental Setup}\label{subsec:microscope}

Due to the size of the eggs of about 0.065 mm$^2$, the analysis of the traps must be done through a microscope. Given this scenario, a setup has been designed to automate image acquisition by taking a total of 165 images per trap over the entire surface of the trap. 

\begin{figure}[htbp]%
    \centering  
    \label{fig:setup}%
        \subfigure[\label{fig:setupv2}]{\includegraphics[width=0.45\columnwidth]{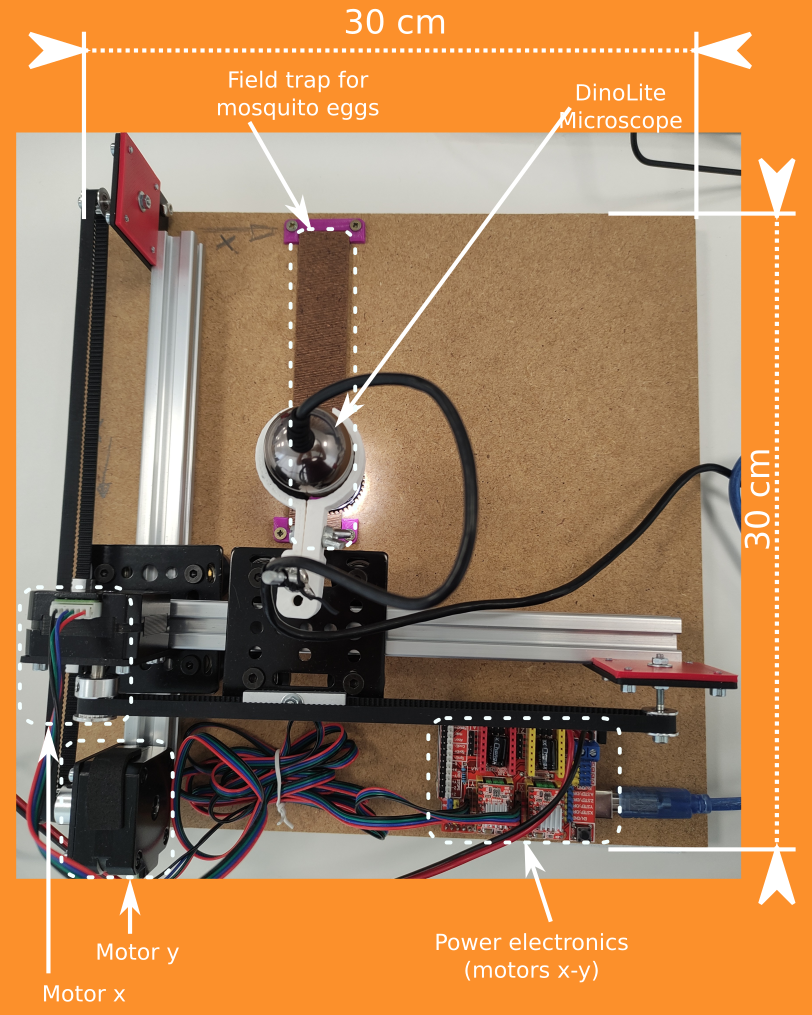}}%
        \hfill
        \subfigure[\label{fig:setuppc}]{\includegraphics[width=0.45\columnwidth]{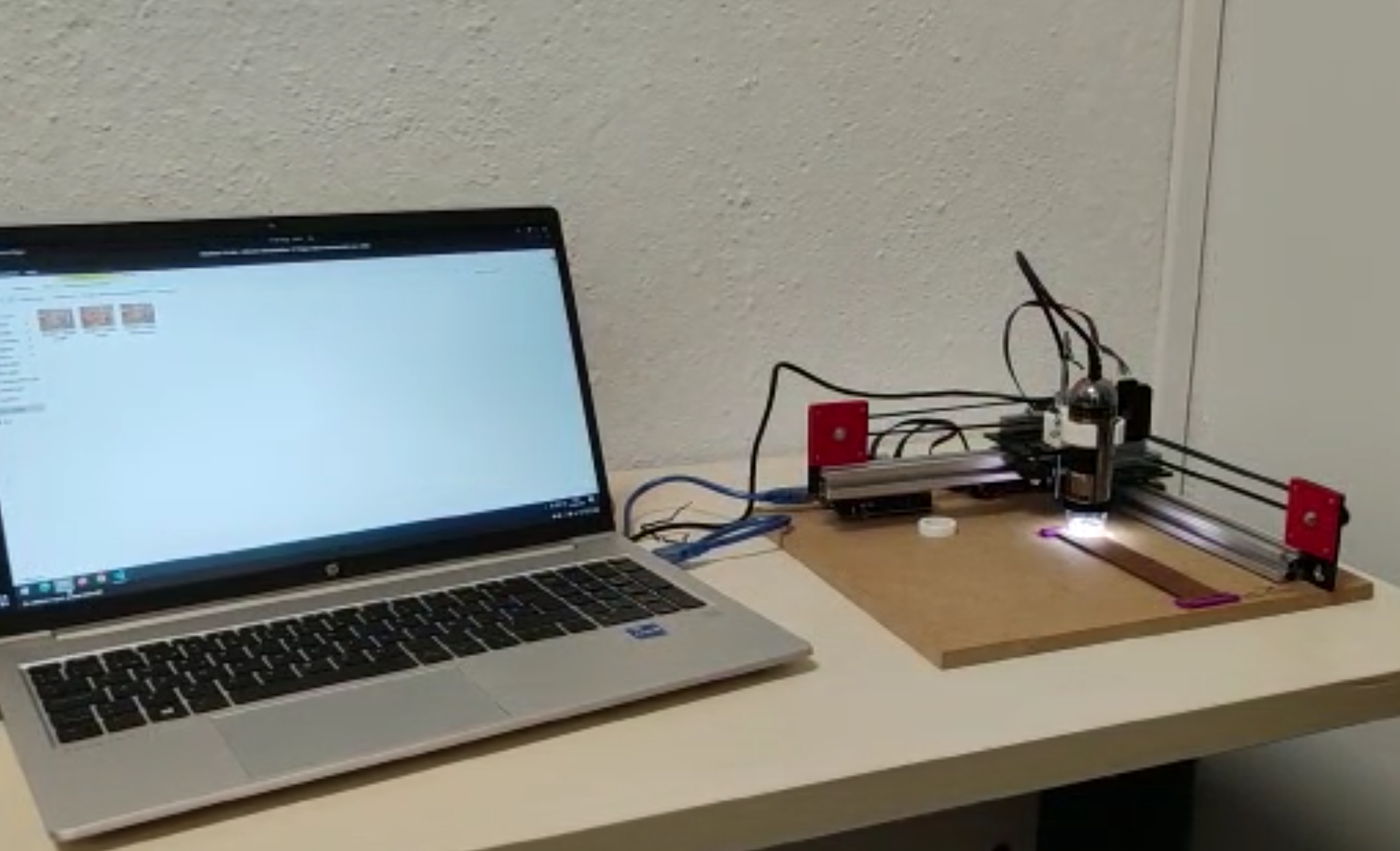}}%
    \caption{(a) A zenithal photo of the experimental setup is shown. In the lower-left corner, the engines in charge of microscope displacement can be appreciated. In the center of the image, the microscope can be seen above the trap. The whole setup is on a 30 $\times$ 30 cm wooden board. (b) Complete setup with the engine connected to the PC to send the movement and image acquisition commands.}
\end{figure}

The experimental setup consists of an x-y positioner composed of two axes\footnote{\url{https://www.turibot.es/maquina-de-dibujo-en-kit}}, two engines, a power electronic board for the movement of the engines and a support for holding the microscope (3D printed piece). The microscope used corresponds to the DinoLite AM4013MZT model\footnote{\url{https://www.dino-lite.eu/es/am4013mzt}} (see Figure~\ref{fig:setupv2}).

The trap dimension is 14.8$\times$2.5 cm. Each image represents about 9$\times$5 mm$^2$. The microscope motion is performed with an overlap of 25\% on the vertical axis (major axis driven by the engine and according to Figure~\ref{fig:setupv2}) and 40\% on the horizontal axis (minor axis according to Figure~\ref{fig:setupv2}). This leads to a total of 33 images per horizontal (along the vertical axis) location and 5 different horizontal locations. Obtaining a total of 165 images per trap. The run over the trap incorporates a 2-second sleep after the motion and before the image acquisition. This holding process is necessary to avoid blurred images resulting from the motion. Thus, the data acquisition time of a trap is 6.33 minutes. The complete setup is shown in Figure~\ref{fig:setuppc}.

\subsection{Object-Segmentation}\label{subsec:segmentation}

Object segmentation \cite{minaee2021image} is a fundamental task in the field of computer vision, involving the identification and delineation of each object present in an image. Unlike image classification, which assigns a single label to an entire image, object segmentation provides precise masks for each instance of an object, enabling a more detailed and granular understanding of the visual content. This capability is essential across a wide range of applications, including autonomous driving, medical imaging, robotics, and augmented reality.

In recent years, advancements in deep learning have revolutionized the field of object segmentation. Among the most prominent models are those from the Mask R-CNN (Region-Based Convolutional Neural Networks) family \cite{He_2017_ICCV}, which have set new standards in terms of accuracy and efficiency. Mask R-CNN is an extension of the Faster R-CNN model, combining object detection with instance segmentation, allowing not only the identification of what objects are present in an image but also the generation of precise segmentation masks for each instance.

The architecture of Mask R-CNN consists of several key components that contribute to its superior performance. First, it utilizes a backbone network, such as ResNet or ResNeXt, to extract rich features from the input images. Next, a region proposal network (RPN) generates candidate regions that might contain objects. These regions are then refined through an additional stage to better adjust the bounding boxes. Finally, a third component, specific to Mask R-CNN, is responsible for predicting segmentation masks for each refined bounding box, enabling pixel-level precision in segmentation.

The adoption of Mask R-CNN and its variants has shown impressive results on standard benchmarks such as COCO (Common Objects in Context) \cite{lin2014microsoft}, where it has consistently outperformed other approaches in instance segmentation tasks. Its versatility also allows for adaptation to various domains and contexts, demonstrating robustness in a variety of practical scenarios.

Despite its success, the development and implementation of Mask R-CNN models are not without challenges. Issues such as high computational cost, the need for large amounts of annotated data, and the fine-tuning of hyperparameters for different applications remain active areas of research. Nevertheless, the benefits it offers in terms of accuracy and detailed segmentation capability make it an invaluable tool in the arsenal of computer vision techniques.

In this paper, Mask R-CNN and Cascade Mask R-CNN are used as a model for the segmentation of the eggs present in the traps. In addition, each instance must be classified as either hatched or unhatched (\textit{full} label is used).

\subsection{Dataset}\label{subsec:dataset}

Image collection was performed at at the facilities of \textit{Conselleria de Agricultura, Ganaderia y Pesca} of the \textit{Generalitat Valenciana}.

The CVAT\footnote{\url{https://www.cvat.ai/}} tool was used for sample labeling. SAM model integration has been used to speed up the process. Consecutive images have not been labeled to avoid labeling the same egg twice and to prevent possible overfitting.

The dataset obtained consisted of 96 images for the training set and 24 for the test set. Table~\ref{tab:splits} shows the number of instances of each class in each set.

\begin{table}[]
    \centering
    \begin{tabular}{lcc}
        \toprule
        Split & Hatch & Full (Unhatch) \\
        \midrule
        Training & 182 & 1042 \\
        Test & 33 & 118 \\
        \bottomrule
    \end{tabular}
    \caption{Performance comparison of different models on test set}
    \label{tab:splits}
\end{table}

\section{Results}\label{sec:results}

\subsection{Segmentation Results}\label{subsec:seg}

The results obtained on the test set can be seen in Table~\ref{tab:model_performance}. The metrics used are  mAP@.5 and mAP@.5:.95 \cite{redmon2018yolov3, lin2014microsoft}

\begin{table}[]
    \centering
    \begin{tabular}{lcc}
        \toprule
        Model & mAP@.5 & mAP@.5:.95 \\
        \midrule
        Mask-RCNN & \textbf{0.91} & 0.66 \\
        Cascade Mask-RCNN & 0.89 & 0.65 \\
        \bottomrule
    \end{tabular}
    \caption{Performance comparison of different models on test set}
    \label{tab:model_performance}
\end{table}

\subsection{Segmentation Examples}\label{subsec:images}

\begin{figure}[h]%
    \centering  
    \label{fig:examples}%
        \subfigure[\label{fig:original}]{\includegraphics[width=0.45\columnwidth]{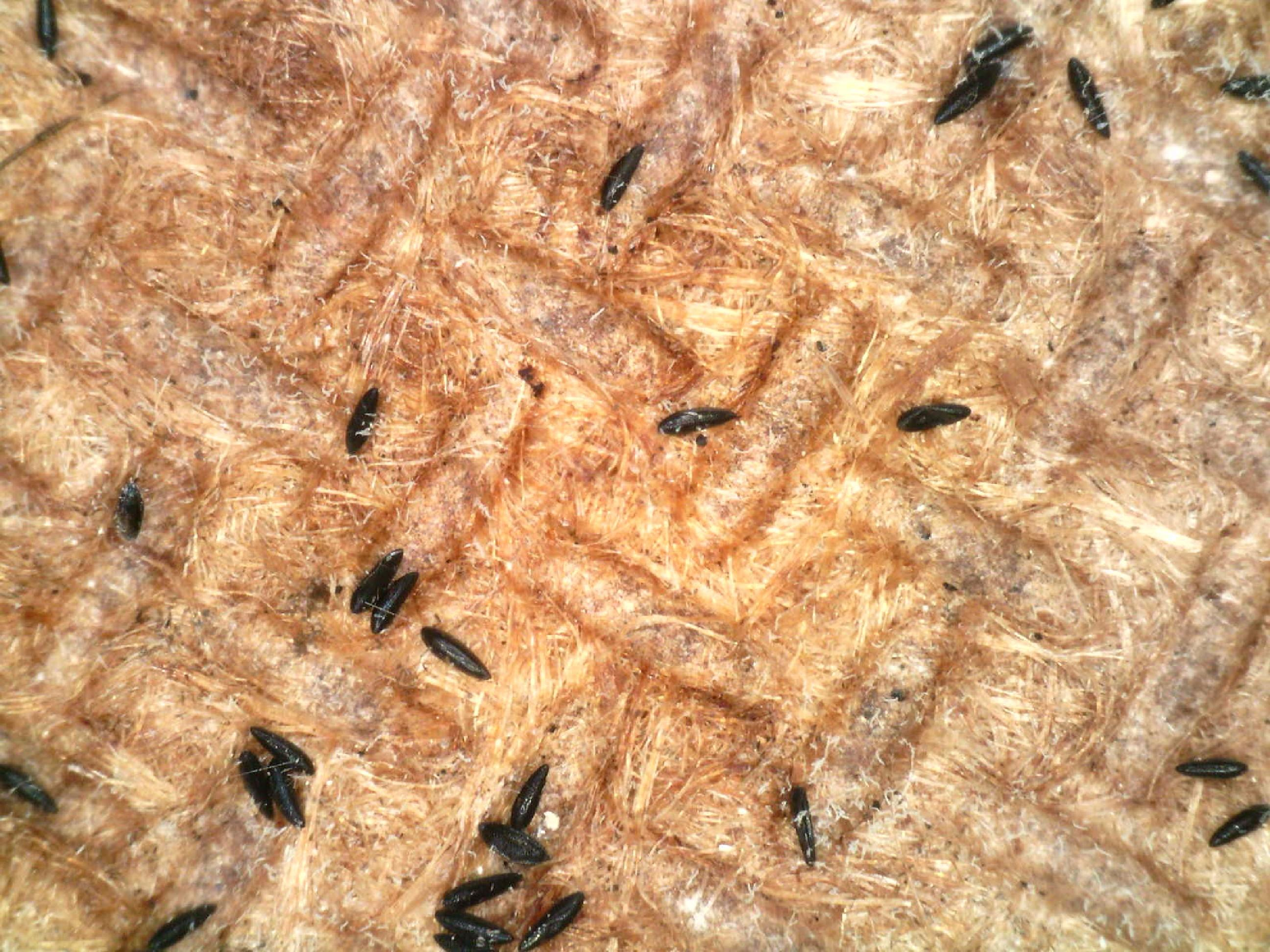}}%
        \hfill
        \subfigure[\label{fig:pred}]{\includegraphics[width=0.45\columnwidth]{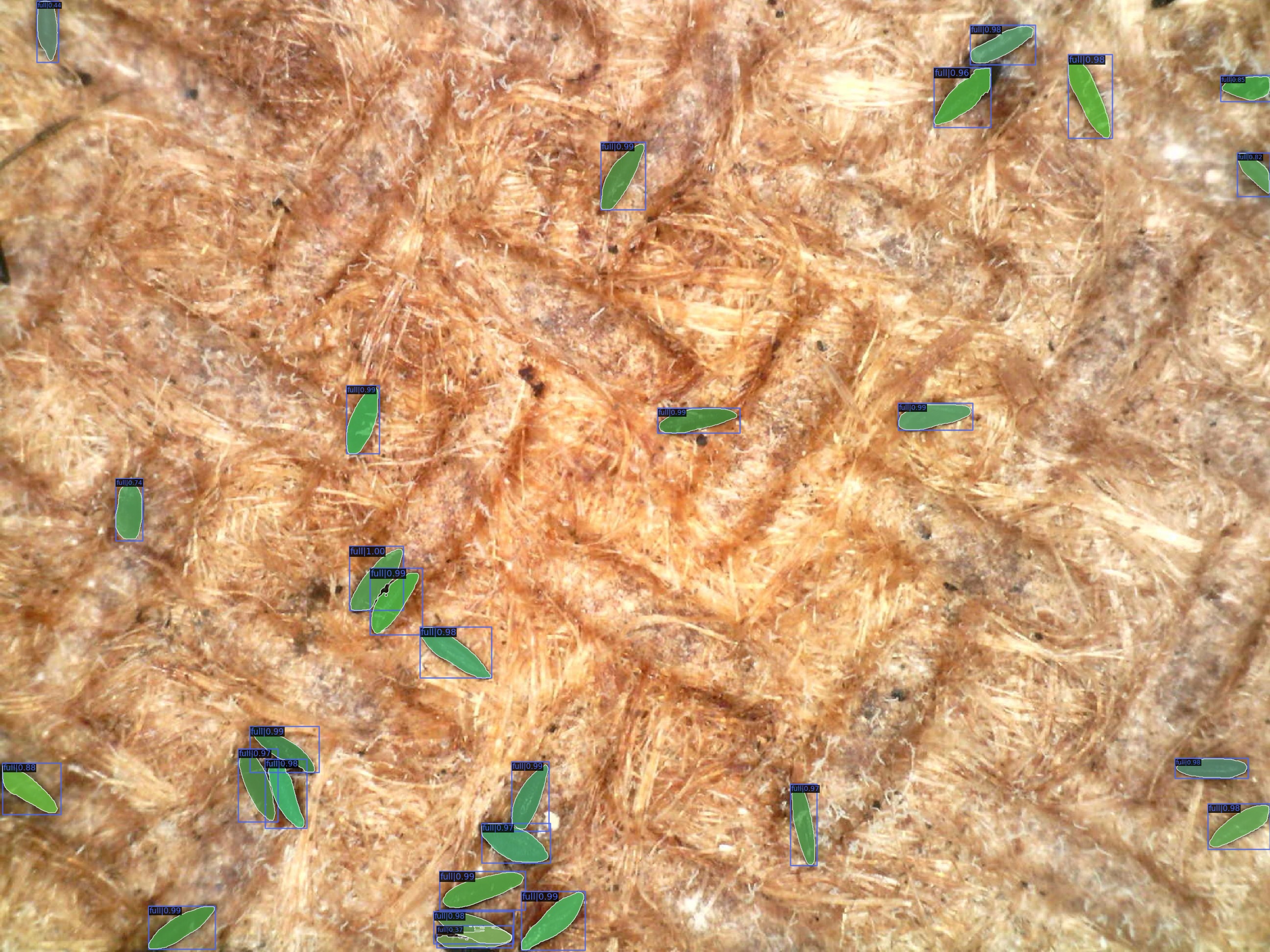}}%
    \caption{(a) Original image taken by the microscope (b) Prediction Mask-RCNN model}
\end{figure}

Figure~\ref{fig:pred} shows an example of egg segmentation on a tablet image. 

\section{Conclusion}\label{sec:conclusion}

This paper presents a complete framework for the analysis of mosquito eggs in field traps using AI models.

The first part of the framework consists of an acquisition setup. The system consists of a microscope and motor-driven axes that move the microscope through the trap. As a result, 165 images per trap are obtained. 

The analysis is performed with a Mask RCNN family model that segments and classifies the eggs into two categories: hatch (hatched eggs) and full (unhatched eggs).

\section{Acknowledgements}\label{sec:ack}

This technical report describes part of the technical work, together with the results obtained, of the project Application of Industry 4.0 techniques to the production of tiger mosquitoes for the Sterile Insect Technique (MoTIA2, file IMDEEA/2022/70), financed by the Valencian Institute for Business Competitiveness (IVACE) and the FEDER funds, executed from July 2022 to June 2023. 

The participation of Javier Naranjo-Alcázar, Jordi Grau-Haro and Pedro Zuccarello in this research has been possible thanks to funding from IVACE and FEDER funds through the MoTIA2 project (IMDEEA/2022/70). The participation of David Almenar in this research has been financed by the Conselleria de Agricultura, Agua, Ganadería y Pesca of the Generalitat Valenciana and the Subdirección de Innovación y Desarrollo de Servicios (TRAGSA group). Jesús López-Ballester has participated in this work in the design of the experimental devices through a subcontract financed with funds from the MoTIA2 project (IMDEEA/2022/70). 

All biological samples and field traps were provided by the \emph{aedes albopictus} mosquito biofactory for the SIT program led and financed by the Conselleria de Agricultura, Agua, Ganadería y Pesca de la Generalitat Valenciana. 



\bibliographystyle{ieeetran}
\bibliography{sample}

\end{document}